\documentclass{article}

\PassOptionsToPackage{numbers, compress}{natbib}

\usepackage[main,nonanonymous]{neurips_2026}

\usepackage[utf8]{inputenc} 
\usepackage[T1]{fontenc}    
\usepackage{hyperref}       
\usepackage{url}            
\usepackage{booktabs}       
\usepackage{amsfonts}       
\usepackage{nicefrac}       
\usepackage{microtype}      
\usepackage{xcolor}         
\newcommand{\cmark}{\ding{51}} 
\newcommand{\xmark}{\ding{55}}
\usepackage{amsmath}
\usepackage{pifont}
\usepackage{graphicx}

\usepackage{subcaption}
\usepackage{wrapfig}

\title{DaCe-DT: Data-Centric Offline Multi-Task Reinforcement Learning via Adaptive Prompts and Trajectory Correction for Heterogeneous Tasks}

\author{%
	Shudong Wang
	\And
	Xinfei Wang
	\And
	Chenhao Zhang
	\And
	Shanchen Pang
	\And
	Wenhao Li
	\And
	Haiyuan Gui
	\And
	Meng Han
	\And
	Xiaojian Liao
}

\begin{document}

	\maketitle
	\nolinenumbers
	
	\begin{abstract}
		Offline multi-task reinforcement learning (Offline MTRL) heavily depends on the quality and distribution of pre-collected data. However, existing methods mainly focus on algorithmic optimization, with less emphasis on data-level improvements to enhance learning ability and generalization performance.
		This paper, from a data perspective, reveals three key bottlenecks that limit Offline MTRL performance: (i) ineffective utilization of prompts length under diverse task complexities, and (ii) semantic irrelevance of randomly sampled prompt segments, (iii) misleading supervision induced by fragmented and discontinuous trajectories.
		To address these challenges, we propose DaCe-DT, a robust offline MTRL framework designed to be insensitive to heterogeneous task complexities and data quality, featuring length-gated prompt masking (LGPM), retrieval-augmented prompt construction (RAPC), and value-adaptive return calibration (VARC). Together, these mechanisms enable DaCe-DT to deliver data-centric prompt adaptation and trajectory refinement, resulting in robust multi-task generalization and stable policy learning amid heterogeneous offline data and tasks. Experimental results on Meta-World show that DaCe-DT consistently outperforms state-of-the-art methods, achieving an average improvement of 11.73\% on optimal datasets and an improvement of 13.34\% on suboptimal datasets, demonstrating its effectiveness in learning stably from imperfect data and improving overall multi-task performance.
	\end{abstract}

	\section{Introduction}
	
	Offline reinforcement learning (RL)~\cite{levine2020offline} enables agents to learn effective policies from static, pre-collected datasets without any active environment interactions. This property offers significant advantages in safety-critical or data-intensive domains such as robotics~\cite{kumar2021workflow}, autonomous driving~\cite{shi2021offline}, and healthcare~\cite{ghasemi2025personalized} by reducing exploration risks and data collection costs. However, traditional offline RL methods typically learn each task in isolation and struggle to generalize their learned knowledge to new tasks. When faced with a novel task~\cite{teh2017distral}, a standard offline RL policy must be retrained from scratch, limiting its scalability in complex multi-task environments. On the other hand, multi-task reinforcement learning (MTRL) addresses cross-task generalization by training a single model on multiple tasks jointly, encouraging the reuse of learned skills and improving data efficiency. Combining these two paradigms, \textbf{offline multi-task reinforcement learning (Offline MTRL)} leverages diverse offline datasets for knowledge transfer without additional interaction, and has attracted growing attention for learning a single versatile policy across tasks. 
	
	Recent advances in sequence modeling provide a promising pathway for offline RL\cite{sun2022paco,yu2020gradient,yang2020multi}. In particular, the Decision Transformer (DT)~\cite{chen2021decision} formulates offline RL as return-conditioned conditional sequence generation over trajectories, achieving impressive results in single-task settings. Building on this idea, prompt-based Decision Transformers\cite{xu2022prompting} extend DT to multi-task settings by prepending task-specific context prompt (e.g., example transitions or a task descriptor) to the input sequence, enabling task inference and task-conditioned behavior. A growing body of work has explored prompt-based DT variants for offline MTRL. These methods address multi-task challenges from different angles: Prompt-DT\cite{xu2022prompting} and Prompt-tuning DT\cite{hu2023prompt} enrich task-relevant context and background evidence through prompt construction; HarmoDT ~\cite{hu2024harmodt} mitigates cross-task gradient conflicts via mask-/subnetwork-based designs; and M3DT ~\cite{kong2025mastering} improves capacity and scalability by leveraging a Mixture-of-Experts (MoE) mechanism to support a larger number of tasks. Despite their successes, most existing efforts predominantly emphasize architectural or algorithmic modifications, while largely overlooking a fundamental data-centric challenge: how to exploit static offline data under pronounced heterogeneity across (and within) tasks. By revisiting these representative works through a data-centric lens, we identify three key bottlenecks that constrain learnability and generalization, motivating our three core observations.

	First, data quality is highly heterogeneous across tasks and even across training mini-batches, and the uniform prompt length limits the model’s generalization and learning efficiency. For example, the default length of the current work is 5. On \textit{Coffee-Push} task, using an empirically tuned “optimal” fixed length (30) improves the single-task success rate by 60\% over the state-of-the-art baseline. Further, with our adaptive prompting strategy, the success rate on \textit{Coffee-push} further increases to 100\%; meanwhile, across all 50 tasks, our approach improves the overall average success rate by \textbf{9.60\%} over the state of the art. (see Observation I, \S \ref{moti1})
	%
	
	
	Second, randomly sampled prompt segments disregard their semantic relevance to the current decision state, limiting effective utilization of heterogeneous offline data. As shown in Fig.~\ref{fig:mot1-2}(c)--(d), RAPC improves the final success rate by 8.4\% over random selection and 9.6\% over fixed selection, while increasing the mean prompt-trajectory similarity from $0.462$ and $0.361$ to $0.741$, respectively. These results show that a static, context-agnostic prompt cannot capture the behavioral evidence most aligned with the current policy state, leaving informative high-quality trajectory segments in the offline dataset underexploited. (see Observation~II, \S\ref{moti2})
	
	Third, existing return-conditioned sequence modeling methods for offline MTRL still lack reliable trajectory stitching capability. Although value learning has been effective for stitching in single-task offline RL~\cite{yamagata2023q}, extending it to offline MTRL is non-trivial. For example, MTIQL~\cite{he2023diffusion} relies on task-specific value functions and stitching modules, introducing considerable overhead and making it difficult to directly apply to prompt-based sequence models without explicit task IDs. Moreover, direct value-based correction under heterogeneous task data may induce biased reachable-return estimates, causing OOD conditioning and false stitching points. (see Observation~III, \S\ref{moti3})

	To address these data-centric bottlenecks in Offline MTRL, we propose DaCe-DT, a unified framework that explicitly adapts context construction and trajectory composition to heterogeneous offline data. DaCe-DT integrates three key mechanisms: \textbf{(1) Length-Gated Prompt Masking (LGPM)}, which learns a task-specific effective prompt length via a differentiable gating parameter, adaptively suppressing redundant or insufficient context for each task; \textbf{(2) Retrieval-Augmented Prompt Construction (RAPC)}, which replaces the static prompt at each step with the most semantically relevant high-return trajectory segment, retrieved via the model’s own GPT-2 embedding layer; and \textbf{(3) Value-Adaptive Return Calibration (VARC)}, which trains a task-conditioned value function with per-task adaptive expectile parameters to calibrate achievable returns at inference, enabling reliable trajectory stitching while avoiding OOD conditioning. 
	
	Beyond RTG calibration, VARC also provides return-level guidance for prompt retrieval by identifying high-quality trajectory segments consistent with the current return target. In DaCe-DT, RAPC retrieves semantically relevant high-return segments, VARC filters them according to achievable-return consistency, and LGPM trims the retrieved prompt to the task-specific effective length. This coordinated design unifies adaptive context construction and reliable trajectory stitching. On the Meta-World benchmark~\cite{yu2020meta}, DaCe-DT improves over the state-of-the-art HarmoDT~\cite{hu2024harmodt} by 11.73\% on the optimal dataset and by 13.34\% on the suboptimal dataset.
	
	In summary, our contributions are as follows:
	\begin{itemize}
		\item We identify, from a data-centric perspective, three key bottlenecks constraining Offline MTRL performance: prompt length mismatch, semantic irrelevance of randomly sampled prompts, and trajectory fragmentation.
		\item We propose three data-centric mechanisms—LGPM, RAPC, and VARC—to address prompt length mismatch, semantic irrelevance, and trajectory fragmentation, respectively.
		\item We conduct extensive experiments on multiple task combinations of the Meta-World benchmark, demonstrating consistent improvements over state-of-the-art methods across datasets of varying quality.
	\end{itemize}

	\section{Preliminary}
	\subsection{Decision Transformer (DT)}
	Decision Transformer (DT)~\cite{chen2021decision} reformulates reinforcement learning as a conditional sequence modeling problem.
	Instead of learning value functions or computing policy gradients, DT trains a causally masked Transformer~\cite{vaswani2017attention} to directly predict actions from offline trajectories.
	Concretely, each trajectory is tokenized as an interleaved sequence of return-to-go, states, and actions,
	$\tau = (\widehat{R}_1, s_1, a_1, \widehat{R}_2, s_2, a_2, \dots, \widehat{R}_T, s_T, a_T)$,
	where the return-to-go is defined as $R_t=\sum_{t'=t}^T r_{t'}$.
	With autoregressive modeling, the policy predicts $a_t$ conditioned on the prefix $\tau_{1:t}$.
	DT is trained using a standard behavior cloning loss:
	\begin{equation}
		\mathcal{L}_{DT} = \mathbb{E}_{\tau \sim \mathcal{D}} \left[ \sum_{t=1}^T \left\| \pi_\theta(\tau_{1:t}) - a_t \right\|_2^2 \right].
	\end{equation}
	The return-to-go token acts as a goal-conditioning signal, enabling goal-directed behavior at test time.

	
	\subsection{Prompt as Task Evidence in Offline MTRL}\label{prompt_func}
	Prompt-based Decision Transformers extend DT to multi-task settings by prepending task-specific context prompts, enabling task inference and task-conditioned behavior. We follow this paradigm and further improve prompt utilization under multi-task heterogeneity.
	
	As illustrated in Fig.~\ref{fig:prompt}, a prompt serves as an explicit task descriptor in the trajectory space, fulfilling three key functions: \ding{172} \textbf{task identification}—enabling a shared Transformer to infer task identity from context alone; \ding{173} \textbf{prior knowledge injection}—encoding task-dependent reward/dynamics patterns and behavioral priors; and \ding{174} \textbf{return conditioning}—specifying goal information to steer the generated trajectory toward desired outcomes.

		\section{\textit{Rethinking Offline MTRL: A Data-Centric Perspective}}
		
		
		\subsection{Prompt Length Mismatch}\label{moti1}
		%
		\begin{figure*}[t]
			\begin{center}
				\centerline
				{\includegraphics[width=\columnwidth]{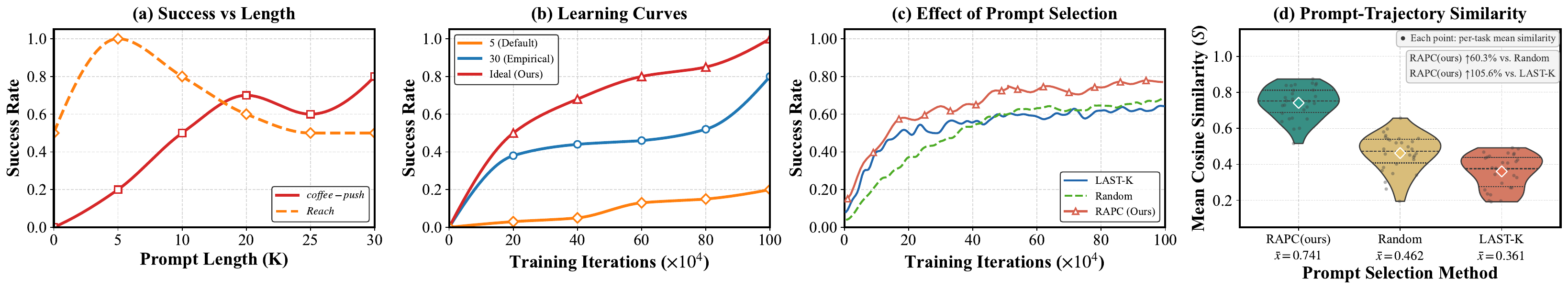}}
				\caption{
					(a) Fixed-length prompt sweep on \textit{Reach} and \textit{Coffee-Push} (MT50) shows each task's "optimal" \textit{K}. (b) On \textit{Coffee-Push}, dynamic prompting outperforms fixed \textit{K}=5 and empirically tuned \textit{K}=30. (c) Comparison of different prompt selection mechanisms in PromptDT on MT30 (curves are smoothed). (d) Cosine similarity between selected prompts and the current inference trajectory under different prompt selection methods.
				}
				\label{fig:mot1-2}
			\end{center}
			\vspace{-14pt}
		\end{figure*}
		\textbf{Observation I:} \textit{Offline MTRL exhibits pervasive data heterogeneity both across tasks and within training mini-batches (different task combinations, see Appendix~\ref{appendix:lgpm_lengths} and Table~\ref{tab:lgpm_lengths}) . Fixed prompt lengths fail to accommodate the divergent contextual requirements of heterogeneous tasks, causing critical under- or over-provisioning of context across different task complexities.}
		
\begin{wrapfigure}{r}{0.52\linewidth}
				\vspace{-10pt}
				\centering
				\begin{subfigure}[t]{0.48\linewidth}
					\centering
					\includegraphics[width=\linewidth]{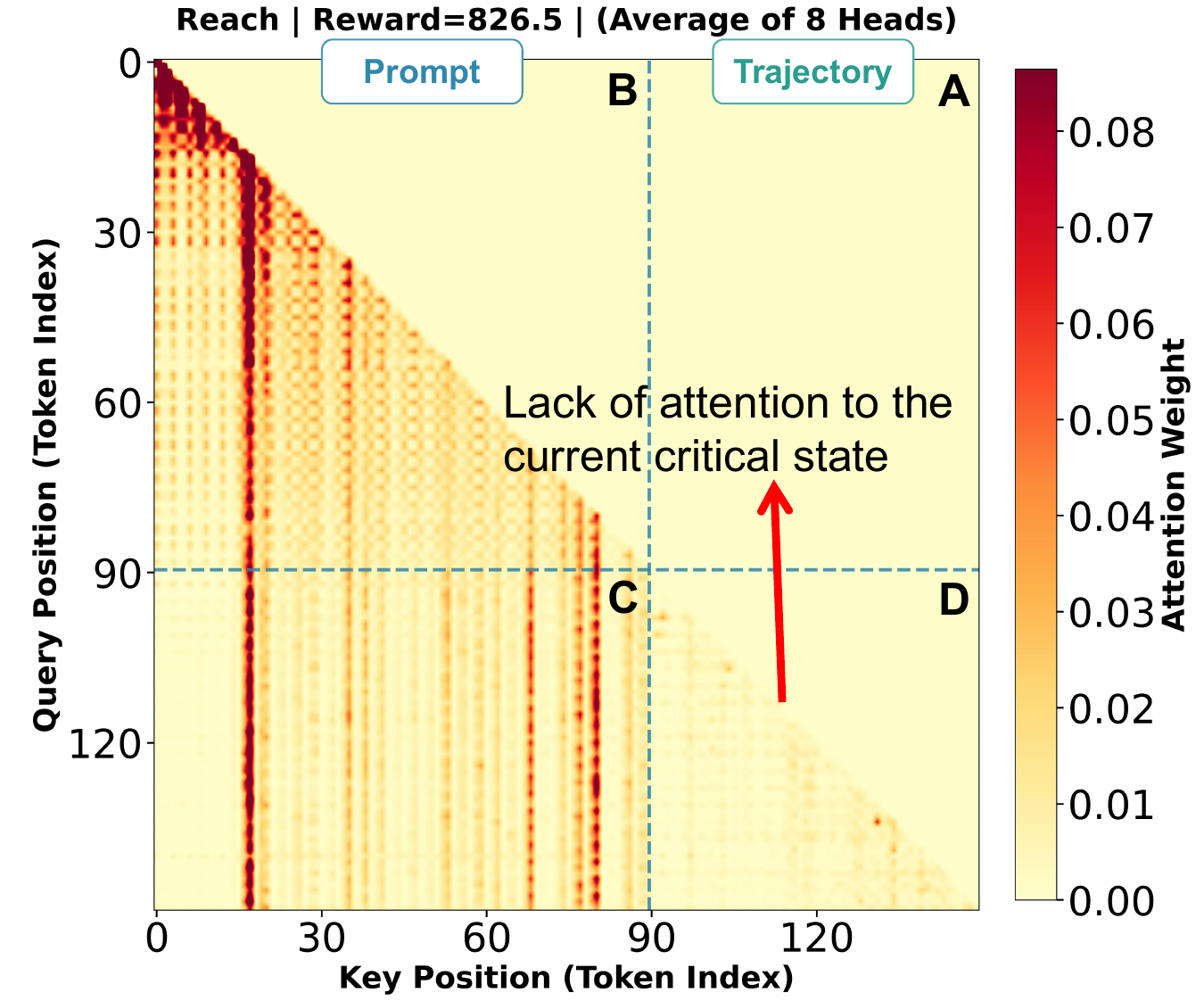}
					\caption*{\textit{Reach}}
					\label{fig:hot:a}
				\end{subfigure}\hfill
				\begin{subfigure}[t]{0.48\linewidth}
					\centering
					\includegraphics[width=\linewidth]{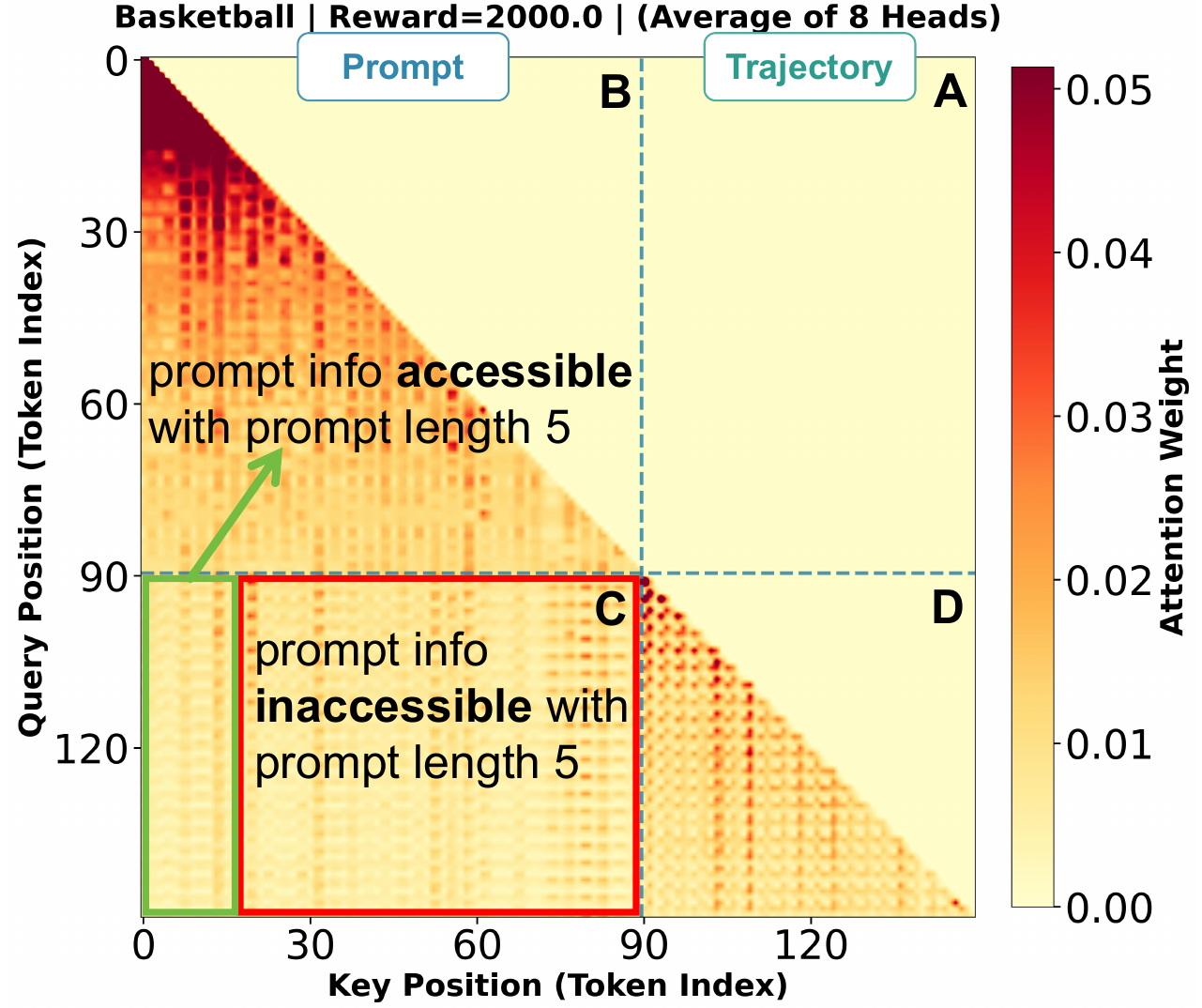}
					\caption*{\textit{Basketball}}
					\label{fig:hot:b}
				\end{subfigure}
				\caption{Task-dependent attention under long prompts ($K{=}30$, $L{=}20$; average over 8 heads). \textit{Reach} (left) shows attention concentrated on recent tokens; \textit{Basketball} relies on broader prompt history.}
				\label{fig:hot}
				\vspace{-4pt}
			\end{wrapfigure}

		As shown in Fig.~\ref{fig:mot1-2}(a), different tasks exhibit markedly different optimal prompt lengths.
		For near-Markovian simple tasks (e.g., \textit{Reach}), a short prompt with $K{=}5$ already achieves a 100\% success rate, while longer prompts introduce redundancy and dilute attention.
		Conversely, for multi-stage complex tasks (e.g., \textit{Coffee-Push}), a longer prompt with $K{=}30$ is required to reach an 80\% success rate, as shorter prompts omit critical historical cues and lead to decision degradation.
		Consequently, no fixed prompt length can simultaneously satisfy the requirements of all tasks.
		Furthermore, Fig.~\ref{fig:mot1-2}(b) demonstrates that even within a single task, a fixed ``optimal'' length fails to accommodate the varying evidence requirements of different trajectory segments during training: prompt lengths that are globally too long or too short at the batch level still induce segment-level mismatches, and adaptive length adjustment yields an additional 20\% improvement over the best fixed baseline (red curve).
		The attention heatmaps in Fig.~\ref{fig:hot} provide a mechanistic explanation: prompts that are too short deprive the model of critical historical information, while prompts that are too long cause attention dilution and cumulative statistical bias.
		It is worth emphasizing that ``adaptive'' here does not refer to searching for a fixed optimal length per task via hyperparameter tuning. Subsequent experiments demonstrate that the optimal prompt length for the same task varies substantially across different task combinations and dataset qualities---a dynamic property that fundamentally precludes any fixed-length search strategy, further underscoring the necessity of a truly adaptive mechanism.

			\subsection{Semantic Irrelevance of Prompts}\label{moti2}
			
			\textbf{Observation II:} \textit{Last-$K$ prompt sampling disregard their semantic relevance to the current decision state, severely limiting the exploitation of informative offline data.}
            
			Even when prompt length is properly addressed, existing methods still suffer from a second, deeper deficiency: they typically first select the highest-return trajectory for each task and then use its last-$K$ segments as prompts~\cite{xu2022prompting,hu2024harmodt,kong2025mastering}. 
            Thus, prompt selection is driven by trajectory-level return and temporal recency, rather than semantic relevance to the current decision state. We further test a random prompt sampling strategy, and the results show that both position-based and random prompt selection ignore the semantic alignment between the prompt and the current decision context.
			
			As shown in Fig.~\ref{fig:mot1-2}(d), the mean cosine similarity between the prompt and the current trajectory is only $0.361$ for fixed prompts and $0.462$ for random prompts, whereas our RAPC raises this to $0.741$---a relative improvement of $105.6\%$ and $60.3\%$, respectively.
			This gap directly translates to training performance: as shown in Fig.~\ref{fig:mot1-2}(c), both fixed and random prompt selection noticeably suppress model performance throughout training, while RAPC---by providing semantically aligned prompts---more effectively exploits the informative structure of the offline dataset and achieves faster convergence with consistently higher success rates.
			These results demonstrate that semantically irrelevant prompt segments are effectively equivalent to noise injection; conversely, retrieving a segment that is semantically aligned with the current state provides the model with precise, stage-level behavioral evidence, substantially improving decision quality.
			
			This motivates the design of RAPC, which leverages the model's own embedding layer to perform semantically-aware prompt retrieval, jointly optimized with LGPM for adaptive context length. As shown in our experiments, introducing RAPC alone yields an improvement of 9.6\% over Prompt-DT.


			\subsection{Sub-Trajectory Fragmentation}\label{moti3}

			\textbf{Observation III:} \textit{Offline data usually only provides partial high-quality sub-trajectories. The existing Offline MTRL methods lack the ability to stitch trajectories, and value-based methods are difficult to be simply applied to offline MTRL scenarios.}
		
\begin{wrapfigure}{r}{0.46\linewidth}
				\vspace{-8pt}
				\centering
				\includegraphics[width=\linewidth]{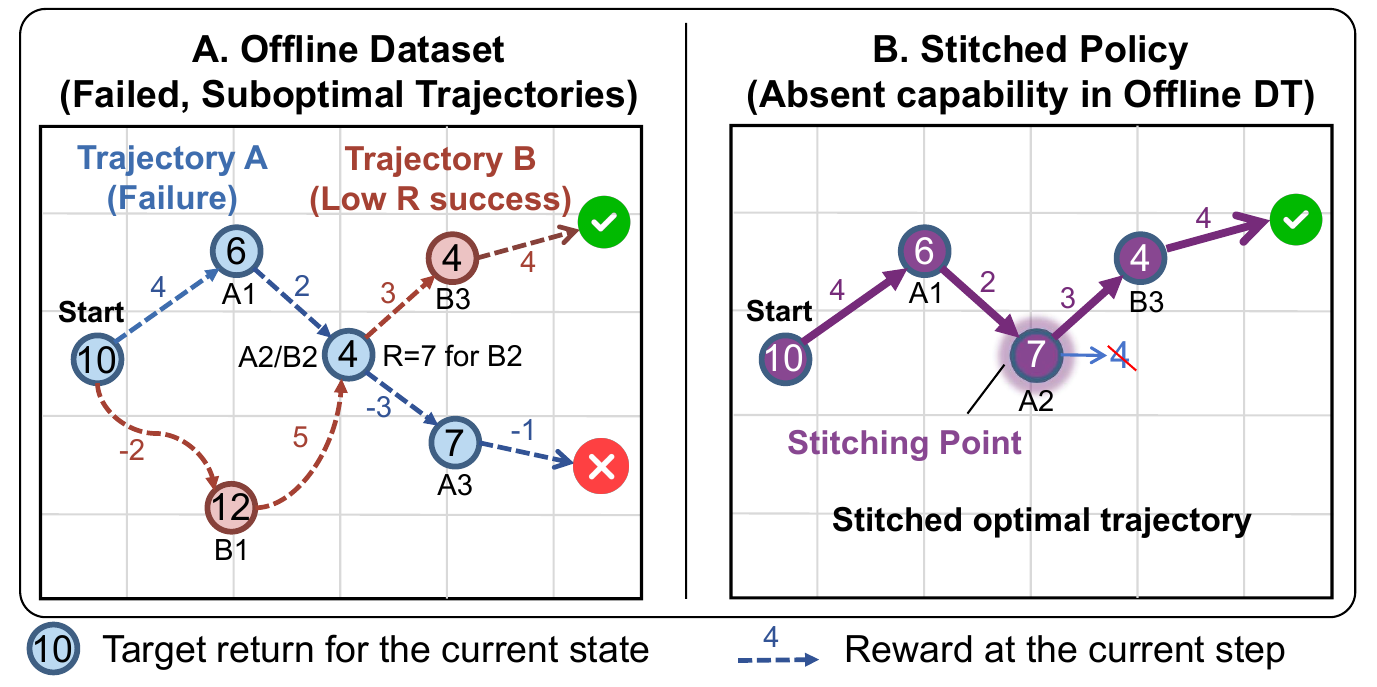}
				\caption{Trajectory stitching and a DT limitation.
					\textbf{Left:} Complementary suboptimal trajectories: $A$ succeeds early but fails later; $B$ is the opposite.
					\textbf{Right:} Optimal stitching ($A_1{\to}A_2{\to}B_3$) is blocked by return-conditioned DTs that cannot adjust RTG at the junction.}
				\label{fig:method}
				\vspace{-6pt}
			\end{wrapfigure}

			Offline datasets often contain complementary locally optimal segments rather than complete globally optimal trajectories, making stitching essential for synthesizing better behaviors.
			As shown in Fig.~\ref{fig:method}, 	Trajectory A provides a high-quality prefix but fails later, while Trajectory B starts suboptimally but ends successfully; ideally, the policy should stitch A's prefix with B's suffix to form a better trajectory. 
			However, return-conditioned DTs are constrained by the empirical support of $(s,\mathrm{RTG})$ pairs: at the stitching state A2, switching to B's successful continuation requires conditioning on a higher target return that is poorly supported in the dataset. 
			Such unsupported RTG conditioning can induce OOD generalization, causing DT to follow the lower-return path and limiting trajectory recombination.
			
			
			In single-task offline RL\cite{yamagata2023q}, value learning has been shown effective for trajectory stitching, and the authors of HarmoDT attempted to extend value correction to IQL for offline MTRL (MTIQL).\cite{he2023diffusion} However, MTIQL relies on task-specific value functions and stitching modules, introducing considerable overhead and weakening the unified modeling advantage of multi-task learning. Moreover, return-conditioned sequence modeling methods usually infer task information from context prompts without explicit task IDs, making such task-specific designs difficult to directly apply. Directly applying value-based correction to these models is also non-trivial: task heterogeneity, uneven data coverage, and different return scales can lead to biased reachable-return estimates, as also observed in our preliminary attempts, causing severe OOD conditioning, unattainable high-return targets, and false stitching points.

				\section{Methodology}
				
\begin{figure*}[t]
					\begin{center}
						\centerline{\includegraphics[width=0.95\textwidth]{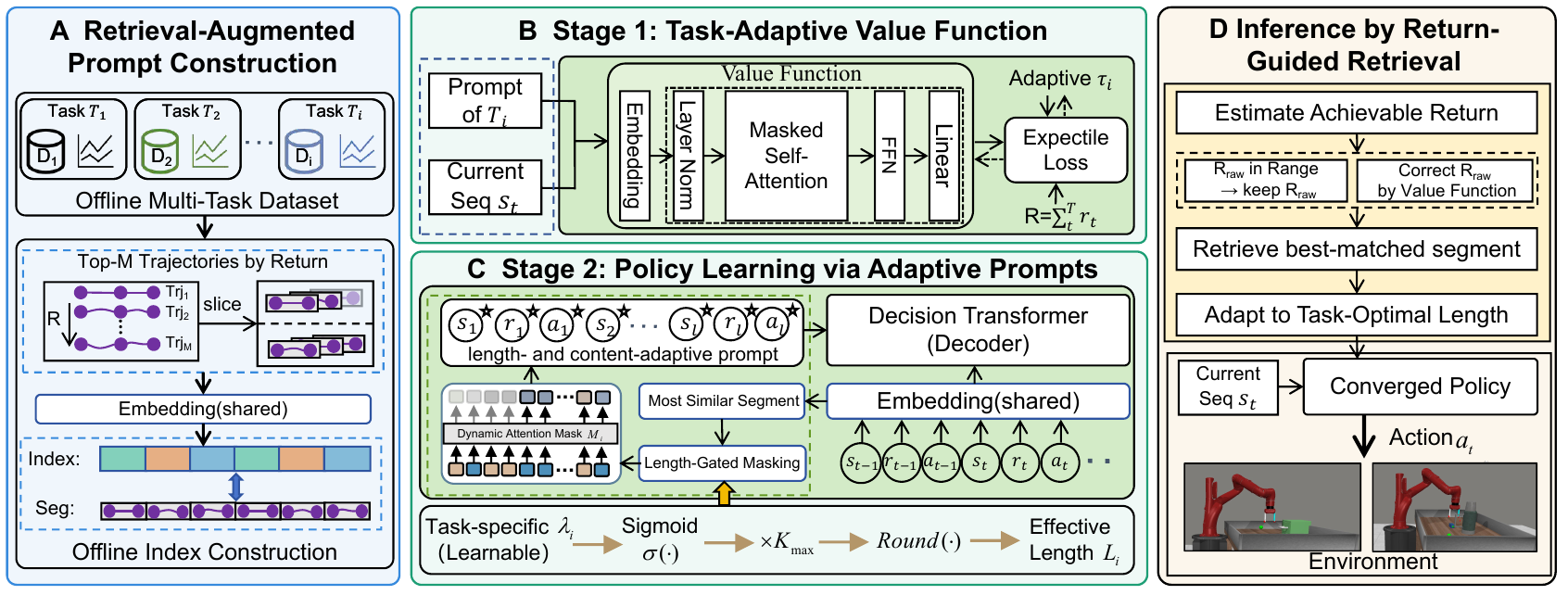}}
						\caption{
                        \textbf{Overview of DaCe-DT.}
(\textbf{A}) RAPC constructs a per-task offline index from top-$M$ high-return trajectory slices via the shared GPT-2 embedding.
(\textbf{B}) Stage~1 trains a task-conditioned value function with per-task adaptive expectile regression.
(\textbf{C}) Stage~2 jointly trains the policy with adaptive prompt length controlled by a dynamic attention mask.
(\textbf{D}) At inference, the value function corrects the return target, which guides retrieval and prompt assembly before the policy outputs $a_t$.
                        }
						\label{framework}
					\end{center}
					\vspace{-12pt}
				\end{figure*}

				In this section, we present \textbf{DaCe-DT}, comprising three core mechanisms: \textbf{I.~Length-Gated Prompt Masking} (LGPM, \S\ref{sec:lgpm_new}), \textbf{II.~Retrieval-Augmented Prompt Construction} (RAPC, \S\ref{sec:rapc_new}), and \textbf{III.~Task-Adaptive Value Function Learning} (VARC, \S\ref{sec:value_new}). The framework is shown in Fig.~\ref{framework}.
				
				\subsection{Length-Gated Prompt Masking (LGPM)}\label{sec:lgpm_new}
				
				To address the issues of \textit{context redundancy} and \textit{information deficiency} caused by fixed prompt lengths in heterogeneous tasks, we propose a length-gated prompt masking mechanism. For each task $T_i$, a learnable scalar $\lambda_i \in \mathbb{R}$ is mapped to a discrete effective length:
				\begin{equation}
					L_i = \mathrm{round}\!\left(\sigma(\lambda_i)\cdot K_{\max}\right), \quad L_i \in \{0,1,\ldots,K_{\max}\}.
				\end{equation}
				A task-specific attention mask $\mathbf{M}_i$ is injected into each Transformer self-attention layer, blocking prompt tokens older than $L_i$ steps while preserving causal order:
				\begin{equation}
					\mathbf{M}_i[j,k] =
						\begin{cases}
							-\infty, & k \in \mathcal{I}_{\text{prompt}} \;\text{and}\; k < (|\mathcal{I}_{\text{prompt}}| - L_i), \\
							-\infty, & k > j, \\
							0, & \text{otherwise}.
					\end{cases}
				\end{equation}
				Since $\mathrm{round}(\cdot)$ is non-differentiable, gradients are passed through it via the Straight-Through Estimator (STE). A sparsity regularization term $\mathcal{L}_{\text{sparse}} = \beta\sum_i \sigma(\lambda_i)$ prevents degeneration to the maximum length, driving each task toward its \textbf{Minimal Sufficient Context}.
				\subsection{Retrieval-Augmented Prompt Construction (RAPC)}\label{sec:rapc_new}
				
				Standard Prompt-DT uniformly samples a fixed demonstration sub-trajectory as the prompt, whose content quality depends on sampling chance with no semantic alignment to the current decision state. We leverage the model’s own GPT-2 embedding layer to build a similarity-based retrieval mechanism, so that at each step the prompt is dynamically retrieved based on the current context.
				
				\textbf{Offline Index Construction.}
				For each task $T_i$, we select the top-$M$ trajectories by episodic return from the offline dataset $\mathcal{D}_i$ (we use $M{=}20$). A sliding window of length $K_q$ steps with stride~1 is applied to yield candidate slices $\{\xi_i^{(j)}\}$. Using only the GPT-2 embedding projection shared with the main policy (without passing through Transformer layers), each step of slice $\xi_i^{(j)}$ is encoded by summing the return-to-go embedding $\phi_R$, state embedding $\phi_s$, action embedding $\phi_a$, and timestep embedding $\phi_{\mathrm{time}}$. The $3K_q$ resulting token vectors are averaged and L2-normalized to obtain the index vector:
				\begin{equation}\label{eq:index}
					\mathbf{e}_i^{(j)} = \mathrm{Norm}\!\left(\frac{1}{3K_q}\sum_{t=1}^{K_q}\!\Bigl[\phi_R\!\left(\hat{R}_t^{(j)}\right) + \phi_s\!\left(s_t^{(j)}\right) + \phi_a\!\left(a_t^{(j)}\right) + \phi_{\mathrm{time}}(t)\Bigr]\right),
				\end{equation}
				where $\mathrm{Norm}(\cdot)$ denotes L2 normalization. All slice vectors, together with source trajectory pointers and slice start indices, are stored in a per-task index $\mathcal{F}_i$.
				
				\textbf{Online Query and Retrieval.}
				At decision step $t$, the query vector $\mathbf{q}_t$ is constructed from the current valid history context using the same encoding procedure as Eq.~\eqref{eq:index}. The most similar slice is retrieved via maximum inner product search (equivalent to cosine similarity under L2 normalization):
				\begin{equation}\label{eq:retrieval}
					j^* = \arg\max_{j\,:\,\mathrm{ret}(\xi_i^{(j)})\,\ge\,\hat{R}_t}\;\mathbf{q}_t^\top \mathbf{e}_i^{(j)},
				\end{equation}
				where $\mathrm{ret}(\xi_i^{(j)})$ denotes the episodic return of the source trajectory of slice $\xi_i^{(j)}$. This return filter ensures the retrieved segment provides behavioral evidence consistent with the current return target, preventing low-quality demonstrations from interfering with stitching decisions.
				
				Since the embedding layer is randomly initialized at the start of training, retrieved similarity carries little semantic meaning early on. We therefore adopt a \emph{progressive retrieval strategy}: static prompts are used for the first $T_{\mathrm{warm}}$ steps; once embeddings have stabilized, the index is refreshed every $\Delta$ steps and retrieval-based prompts are used thereafter.
				
				\subsection{Value-Adaptive Return Calibration (VARC)}\label{sec:value_new}
				
				To enable return calibration at inference time, we train a value function that shares the same GPT-2 architecture as the main policy. Its input format is identical to the policy: the prompt $\mathcal{P}_i$ and the current trajectory history are concatenated and fed into the Transformer, and the model outputs a \emph{scalar estimate of the achievable return} from the current context.
				
				To accommodate heterogeneous data distributions across tasks, for each task $T_i$ we introduce a learnable scalar $\eta_i \in \mathbb{R}$ that parameterizes a task-specific expectile:
				\begin{equation}
					\tau_i = 0.5 + 0.5\cdot\sigma(\eta_i), \quad \tau_i \in (0.5,\,1).
				\end{equation}
				Tasks with high data noise converge toward conservative quantiles ($\tau_i \to 0.5$), while low-noise tasks adopt more aggressive ones ($\tau_i \to 1$). We jointly optimize the shared value parameters $\phi$ and all $\{\eta_i\}_{i=1}^N$ by minimizing the expectile regression loss across all tasks:
				\begin{equation}
					\min_{\phi,\,\{\eta_i\}} \sum_{i=1}^{N} \mathbb{E}_{(\mathcal{H}_t,r,\mathcal{H}_{t+1})\sim\mathcal{D}_i} \Bigl[\bigl|\tau_i - \mathbf{1}(\delta_i < 0)\bigr|\cdot\delta_i^2\Bigr],
				\end{equation}
				where $\delta_i = r + \gamma V_{\phi}(\mathcal{H}_{t+1},\mathcal{P}_i) - V_{\phi}(\mathcal{H}_t,\mathcal{P}_i)$ is the TD error. After convergence, $V_{\phi^*}$ and $\{\tau_i^*\}$ are frozen and used solely at inference time.
				
				\textbf{Policy Training.}
				With the value function frozen, we jointly train the policy $\pi_\theta$ and the task-specific prompt-length parameters $\{\lambda_i\}$ (introduced in \S\ref{sec:lgpm_new}) by minimizing the action prediction loss with sparsity regularization:
				\begin{equation}
					\begin{gathered}
						\min_{\theta,\{\lambda_i\}} \sum_{i=1}^{N}\;
							\mathbb{E}_{\xi \sim \mathcal{D}_i,\, t}\Bigl[
							-\log \pi_\theta\!\left(a_t \mid \mathcal{H}_{t,i}\right)
							\Bigr]
							+ \mathcal{L}_{\text{sparse}},\\
						\mathcal{H}_{t,i}=\bigl(\mathbf{s}_{\le t},\, \mathbf{a}_{<t},\, \mathrm{RTG}_t,\, \mathcal{P}_i,\, \mathbf{M}_i\bigr),\quad \mathcal{L}_{\text{sparse}}=\beta\sum_{i=1}^{N}\sigma(\lambda_i).
					\end{gathered}
				\end{equation}
				The value function is not applied during training; it is used exclusively at inference time to handle out-of-distribution return conditions.
				
				\subsection{Inference: Return Correction, Retrieval, and Length-Adaptive Prompt Assembly}\label{sec:inference_new}
				
				At inference time, the three mechanisms operate in sequence at every decision step.
				
				\textbf{Step~1: Return Correction.}
				The frozen value function $V_{\phi^*}$ checks whether the current raw return target lies within the achievable range. For task $T_i$, given the current trajectory sequence $\mathcal{H}_t = (\mathbf{s}_{\le t}, \mathbf{a}_{<t}, \mathrm{RTG}_t)$:
				\begin{equation}
					\hat{R}_t =
						\begin{cases}
							V_{\phi^*}(\mathcal{H}_t,\,\mathcal{P}_{i,t-1}), & \left|R_t^{\mathrm{raw}} - V_{\phi^*}(\mathcal{H}_t,\,\mathcal{P}_{i,t-1})\right| > \varepsilon, \\[4pt]
							R_t^{\mathrm{raw}}, & \text{otherwise},
					\end{cases}
				\end{equation}
				where $R_t^{\mathrm{raw}}$ is initialized as $V_{\phi^*}(\mathcal{H}_0,\mathcal{P}_{i,0})$ and updated as $R_t^{\mathrm{raw}} = \hat{R}_{t-1} - r_{t-1}$, and $\varepsilon$ is a tolerance threshold that prevents over-correction. When the value function detects a higher achievable return from the current sequence $\mathcal{H}_t$, the corrected $\hat{R}_t$ is raised, directing the policy toward superior continuations and enabling cross-trajectory stitching.
				
				\textbf{Step~2: Retrieval.}
				Using the corrected return $\hat{R}_t$ and the current trajectory history as the query, similarity-based retrieval is performed per Eq.~\eqref{eq:retrieval} to obtain the optimal slice $\xi_i^{(j^*)}$ of length $K_q$ steps.
				
				\textbf{Step~3: Length-Adaptive Prompt Assembly.}
				The retrieved slice is cropped or extended according to the task-optimal length $L_i^*$ learned by LGPM (\S\ref{sec:lgpm_new}):
				\begin{equation}
					\mathcal{P}^*_{i,t} =
						\begin{cases}
							\xi_i^{(j^*)}[K_q - L_i^* + 1\;:\;K_q], & L_i^* \le K_q, \\[4pt]
							\xi_i^{(j^*)}[1:K_q]\;\|\;\xi_i^{(j^*),\mathrm{ext}}[K_q{+}1:L_i^*], & L_i^* > K_q,
					\end{cases}
				\end{equation}
				where $\|$ denotes sequence concatenation and $\xi_i^{(j^*),\mathrm{ext}}$ is the extension from the continuation of the same source trajectory. When $L_i^* \le K_q$, the \emph{last} $L_i^*$ steps of the retrieved slice are retained (temporally closest to the current state); when $L_i^* > K_q$, the full $K_q$-step slice is appended with $L_i^* - K_q$ subsequent steps. The assembled prompt $\mathcal{P}^*_{i,t}$ fully replaces the current prompt tokens fed to the Transformer, ensuring the policy simultaneously receives a calibrated return condition and a contextually aligned behavioral reference at every step.
				
				
				

				\section{Experiments}
				\begin{table}[t]
					\caption{Performance comparison on Meta-World with 5, 30, and 50 randomly sampled tasks under near-optimal and sub-optimal datasets. Each result is averaged over three random seeds. To ensure fair comparison, we adopt the same random seeds as those used in the previous baseline methods.}
					\centering
					\resizebox{\linewidth}{!}{
						\begin{tabular}{l|cc|cc|cc}
							\toprule[2pt]
							Method & \multicolumn{2}{c|}{Meta-World 5 Tasks} & \multicolumn{2}{c|}{Meta-World 30 Tasks} & \multicolumn{2}{c}{Meta-World 50 Tasks} \\
							\cmidrule{2-7}
							& Near-optimal & Sub-optimal & Near-optimal & Sub-optimal & Near-optimal & Sub-optimal \\
							\midrule
							MTDIFF  & $\textbf{100.0} \pm 0.0$ & $66.30 \pm 2.31$ & $67.52 \pm 0.35$ & $54.21 \pm 1.10$ & $61.32 \pm 0.89$ & $48.94 \pm 0.95$ \\
							MTDT & $\textbf{100.0} \pm 0.0$ & $64.67 \pm 5.25$ & $71.89 \pm 0.95$ & $49.33 \pm 2.05$ & $65.80 \pm 1.02$ & $42.33 \pm 1.89$ \\
							Prompt-DT*  & $\textbf{100.0} \pm 0.0$ & $67.00 \pm 2.33$ & $71.30 \pm 0.67$ & $54.33 \pm 0.78$ & $71.60 \pm 1.40$ & $51.40 \pm 0.35$ \\
							HarmoDT-R*  & $\textbf{100.0} \pm 0.0$ & $68.00 \pm 2.33$ & $80.10 \pm 3.12$ & $59.00 \pm 3.66$ & $74.24 \pm 2.66$ & $52.45 \pm 1.17$ \\
							HarmoDT-M*  & $\textbf{100.0} \pm 0.0$ & $72.12 \pm 3.48$ & $79.67 \pm 1.46$ & $\textbf{62.33} \pm 0.66$ & $\textbf{78.80} \pm 0.67$ & $56.20 \pm 0.56$ \\
							HarmoDT-F*  & $\textbf{100.0} \pm 0.0$ & $\textbf{75.04} \pm 2.33$ & $\textbf{81.60} \pm 0.78$ & $60.60 \pm 1.12$ & $77.80 \pm 0.66$ & $ \textbf{57.50} \pm 0.48$ \\
							\midrule
							\textbf{DaCe-DT(Ours)}  & $\textbf{100.0} \pm 0.0$ & $\textbf{88.00} \pm 2.56$ & $\textbf{93.33} \pm 1.67$ & $\textbf{75.67} \pm 1.33$ & $\textbf{88.40} \pm 2.67$ & $\textbf{64.00} \pm 3.45$ \\
							\bottomrule[2pt]
							
					\end{tabular}}
					\vspace{-10pt}
					\label{tb:scale}
				\end{table}
					
				\label{sec:experiments}
				In this section, we conduct comprehensive experiments on the Meta-World~\cite{yu2020meta} benchmark (covering MT5, MT30, and MT50 configurations) to answer the following key research questions:
				\textbf{RQ1 (Comparative Performance):} Does our proposed \textbf{DaCe-DT} outperform state-of-the-art multi-task offline RL baselines across datasets with varying qualities (e.g., near-optimal vs. sub-optimal)?\begin{wraptable}[21]{r}{0.5\linewidth}
					\centering
					\caption{Average success rate across 3 seeds on MT50 with random goals (MT50-rand) under both near-optimal and sub-optimal cases.}
					\vspace{0.1cm}
					\small
					\begin{tabular}{l|cc}
						\toprule[2pt]
						Method & Near-opt. & Sub-opt. \\
						\midrule
						\textbf{CARE}~(online) & $46.12_{\pm 1.30}$ & -- \\
						\textbf{PaCo}~(online) & $54.31_{\pm 1.32}$ & -- \\
						\textbf{Soft-M}~(online) & $53.41_{\pm 0.72}$ & -- \\
						\textbf{D2R}~(online) & $63.53_{\pm 1.22}$ & -- \\
						\textbf{MOORE}~(online) & $72.90_{\pm 3.30}$ & -- \\
						\textbf{ARS-LoRA}~(online) & $\textbf{83.20}_{\pm 1.80}$ & -- \\
						\midrule
						\textbf{MTBC}  & $60.39_{\pm 0.86}$ & $34.53_{\pm 1.25}$ \\
						\textbf{MTIQL} & $56.21_{\pm 1.39}$ & $43.28_{\pm 0.90}$ \\
						\textbf{MTDIFF-P} & $59.53_{\pm 1.12}$ & $48.67_{\pm 1.32}$ \\
						\textbf{MTDIFF-P-ONEHOT} & $61.32_{\pm 0.89}$ & $48.94_{\pm 0.95}$ \\
						\textbf{MTDT} & $65.80_{\pm 1.02}$ & $42.33_{\pm 1.89}$ \\
						\textbf{Prompt-DT*} & $71.60_{\pm 1.40}$ & $51.40_{\pm 0.35}$ \\
						\textbf{HarmoDT-R*} & $74.24_{\pm 2.66}$ & $52.45_{\pm 1.17}$ \\
						\textbf{HarmoDT-M*} & $\textbf{78.80}_{\pm 0.67}$ & $56.20_{\pm 0.56}$ \\
						\textbf{HarmoDT-F*} & $77.80_{\pm 0.66}$ & $\textbf{57.50}_{\pm 0.48}$ \\
						\midrule
						\textbf{DaCe-DT(ours)} & $\textbf{88.40}_{\pm 1.67}$ & $\textbf{64.00}_{\pm 3.45}$ \\
						\bottomrule[2pt]
					\end{tabular}
					\label{tb:50task}
				\end{wraptable}
				\textbf{RQ2 (Generalization \& Robustness):} How does DaCe-DT generalize to unseen tasks, and how robust is it in sparse-reward environments?
				\textbf{RQ3 (Ablation Study):} What are the individual contributions of key design components, such as our LGPM, RAPC and VARC, to the overall performance?

				\subsection{Environment and Baselines}

				\textbf{Environment.}
				We evaluate on the Meta-World multi-task robotic manipulation benchmark, which consists of 50 manipulation tasks sharing the same robot dynamics but differing in objects and interaction modes.
				Following prior work~\cite{hu2024harmodt,he2023diffusion}, we use random-goal evaluation to measure generalization across tasks.
				Performance is reported as the average success rate over all tasks.
				
				\textbf{Offline datasets.}
				We consider two offline training regimes.
				(i) \textit{Near-optimal:} a high-quality dataset generated by SAC-Replay~\cite{haarnoja2018soft} that mixes expert and random trajectories.
				(ii) \textit{Sub-optimal:} a more challenging dataset constructed from early-stage trajectories while retaining only 50\% of expert demonstrations, mimicking realistic imperfect logging.
				
			\begin{wrapfigure}{r}{0.52\linewidth}
				\centering
				\includegraphics[width=\linewidth]{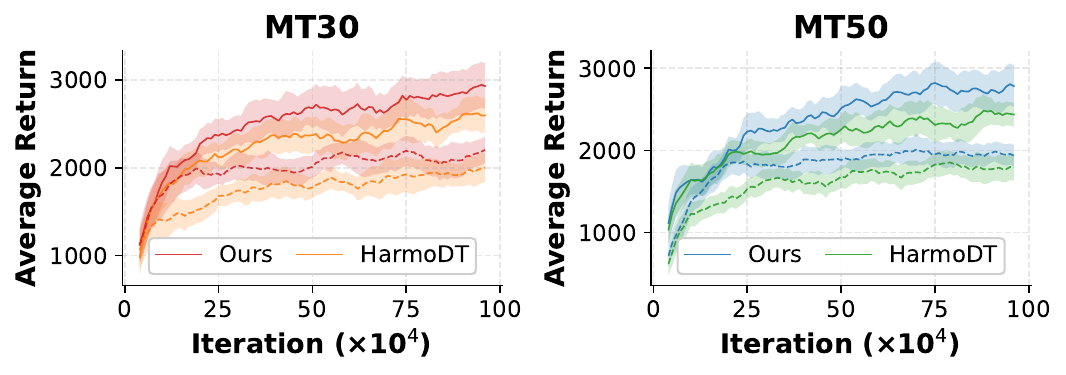}
				\caption{Comparison of training reward trajectories between DaCe-DT and HarmoDT (current offline MTRL state-of-the-art) across MT30 and MT50 configurations of Meta-World.}
				\label{fig:return}
				\vspace{-10pt}
			\end{wrapfigure}
				\textbf{Baselines.} We compare our method against several representative multi-task offline RL baselines:
				(1) \textbf{MTBC}~\cite{he2023diffusion}: Multi-task Behavior Cloning with task ID conditioning.
				(2) \textbf{MTIQL}: Multi-task IQL~\cite{kostrikov2021offline} with multi-head critic and task-conditioned policy network.
				(3) \textbf{MTDIFF-P}~\cite{he2023diffusion}: A diffusion-based policy learning method with prompting and transformer modules.
				(4) \textbf{MTDT}~\cite{he2023diffusion}: Decision Transformer adapted for multi-task learning with task ID embedding.
				(5) \textbf{Prompt-DT}~\cite{xu2022prompting}: Decision Transformer with trajectory prompt, as discussed in \S~\ref{promptdt}.
				(6) \textbf{HarmoDT}~\cite{hu2024harmodt}: Prompt-DT extended with task-specific harmony subspace masks to mitigate cross-task gradient conflicts.
				(7) \textbf{CARE}~\cite{sodhani2021multi}: Uses meta-data and encoder mixing for task representation.
				(8) \textbf{PaCo}~\cite{sun2022paco}: Adopts parameter composition for task-specific parameter reassembly.
				(9) \textbf{Soft-M}~\cite{yang2020multi}: Uses routing networks for modular soft composition.
				(10) \textbf{D2R}~\cite{he2024not}: Employs diverse routing paths for different tasks.
				(11) \textbf{MOORE}~\cite{hendawy2023multi}: Online MTRL method using mixture of orthogonal experts to maintain diverse shared representations.
				(12) \textbf{ARS-LoRA}~\cite{cho2025ars}: Online MTRL method with adaptive reward scaling and LoRA-based parameter-efficient task adaptation.
				We use results from ~\cite{he2023diffusion} for most baselines. * indicates baselines of our own implementation.

				\subsection{Performance Comparison}
					\begin{wraptable}{r}{0.55\linewidth}
					\vspace{-12pt}
					\centering
					\setlength{\tabcolsep}{2pt}
					\caption{Generalization to unseen tasks. Methods trained on a task subset and evaluated without fine-tuning.}
					\label{tab:unseen}
					\vspace{2pt}
					\scriptsize
					\begin{tabular}{@{}lcccc@{}}
						\toprule
						Setting & MTDT & Prompt-DT & HarmoDT-F & \textbf{Ours} \\
						\midrule
						Cheetah-dir & $662.4_{\pm1.3}$ & $935.3_{\pm2.6}$ & $958.5_{\pm1.5}$ & $\mathbf{966.9_{\pm2.2}}$ \\
						Cheetah-vel & $-170.1_{\pm5.7}$ & $-127.7_{\pm9.9}$ & $-66.5_{\pm1.2}$ & $\mathbf{-22.0_{\pm1.2}}$ \\
						Ant-dir     & $165.3_{\pm0.4}$ & $278.7_{\pm38.7}$ & $298.3_{\pm1.0}$ & $\mathbf{355.9_{\pm1.7}}$ \\
						\midrule
						Average     & $219.2$ & $362.1$ & $396.8$ & $\mathbf{433.6}_{9.27\%\uparrow}$ \\
						\bottomrule
					\end{tabular}
				\end{wraptable}
				We report performance comparisons with mainstream baselines on the MT5, MT30, and MT50 benchmarks.
				As shown in Tables~\ref{tb:scale},~\ref{tb:50task} and Fig~\ref{fig:return}, our method demonstrates consistent improvements.
				Specifically, on the near-optimal MT5 dataset, it achieves a success rate of 100\%,
				and on the sub-optimal dataset, it yields a performance gain of 12.96\%.
				For MT30, our method achieves improvements of 11.73\% and 13.34\% on the near-optimal and sub-optimal datasets, respectively.
				For MT50, the improvements are 9.60\% and 6.50\%, respectively.

				\subsection{More Analysis}
				
				\paragraph{Generalization to Unseen Tasks.}
				To evaluate zero-shot generalization, we test DaCe-DT on three unseen continuous-control environments (\textsc{Cheetah-dir}, \textsc{Cheetah-vel}, \textsc{Ant-dir}) following HarmoDT's protocol. The retrieval index is built from a small set of expert segments using frozen embeddings, with prompt length set to the average $L_i^*$ from training tasks. As shown in Table~\ref{tab:unseen}, DaCe-DT outperforms all baselines by 9.27\%, demonstrating that the retrieval index generalizes to unseen tasks at zero training cost.

				\paragraph{Robustness under Sparse Reward.}
				
				To isolate the contribution of RAPC and LGPM, we disable the value function and evaluate on eight custom \textsc{Maze2D} layouts with entirely sparse rewards.
				As shown in Table~\ref{tab:maze2d}, our method achieves an average score of 36.9 vs.\ 18.2 for Prompt-DT, demonstrating that embedding-based retrieval alone substantially improves prompt quality without any return calibration.

				\subsection{Ablation Study}
				
				\paragraph{Component Ablation.}
					\begin{wraptable}[12]{r}{0.50\linewidth}
    \centering
    \vspace{-12pt}
    \setlength{\tabcolsep}{4pt}
    \caption{\textbf{Ablation on MT30.} Impact of LGPM, RAPC and VARC on Success Rate and Return.}
    \label{tab:ablation_mt30}
    \begin{tabular}{ccccc}
        \toprule
        \multicolumn{3}{c}{\textbf{Components}} & \multicolumn{2}{c}{\textbf{MT30}} \\
        \cmidrule(r){1-3} \cmidrule(l){4-5}
        LGPM & RAPC & VARC & Succ. & Return \\
        \cmidrule(r){1-3} \cmidrule(l){4-5}
        \xmark & \xmark & \xmark & 71.3 & 2436.8 \\
        \cmark & \xmark & \xmark & 83.7 & 2864.5 \\
        \xmark & \cmark & \xmark & 85.3 & 2961.4 \\
        \xmark & \xmark & \cmark & 82.6 & 2735.6 \\
        \cmark & \cmark & \cmark & \textbf{93.3} & \textbf{3185.5} \\
        \bottomrule
    \end{tabular}
\end{wraptable}
               We evaluate the average success rate of model variants by ablating or replacing key
components under the MT30:
\textbf{(1) With LGPM only}: The dynamic attention mask is retained to learn
task-adaptive prompt lengths, but retrieval-augmented prompt construction (RAPC)
and return calibration (VARC) are both disabled.
\textbf{(2) With RAPC only}: Retrieval-augmented prompt construction is enabled,
but task-adaptive prompt length (LGPM) and return calibration (VARC) are both disabled.
\textbf{(3) With VARC only}: Return calibration is applied at inference, but
prompts fall back to the fixed sub-trajectory baseline. Results are shown in Table~\ref{tab:ablation_mt30}.
				
				\section{Conclusion}
				
				In summary, we identify three data-centric bottlenecks limiting learnability and generalization in Offline MTRL: (i) prompt length mismatch under heterogeneous task complexities, (ii) semantic irrelevance of randomly sampled prompts, and (iii) misleading supervision from fragmented trajectories. We address them with \textbf{DaCe-DT}, which integrates length-gated prompt masking (LGPM) for adaptive context selection, retrieval-augmented prompt construction (RAPC) for semantically aligned behavioral evidence, and value-adaptive return calibration (VARC) for reliable trajectory stitching at inference. Extensive experiments on Meta-World demonstrate consistent and significant improvements over state-of-the-art methods across datasets of varying quality.
				
				\begin{ack}
					Use unnumbered first level headings for the acknowledgments. All acknowledgments
					go at the end of the paper before the list of references. Moreover, you are required to declare
					funding (financial activities supporting the submitted work) and competing interests (related financial activities outside the submitted work).
					More information about this disclosure can be found at: \url{https://neurips.cc/Conferences/2026/PaperInformation/FundingDisclosure}.

					Do {\bf not} include this section in the anonymized submission, only in the final paper. You can use the \texttt{ack} environment provided in the style file to automatically hide this section in the anonymized submission.
				\end{ack}
				

				%
				%
					%
					%
					%
					%
					%
					%
				\bibliographystyle{plain}
				\bibliography{example_paper}

				
				\appendix
				
				\section{Detailed Environment and Baselines}
				\label{appendix:baselines}
\begin{figure*}[t]
					\centering
					\includegraphics[width=0.98\textwidth]{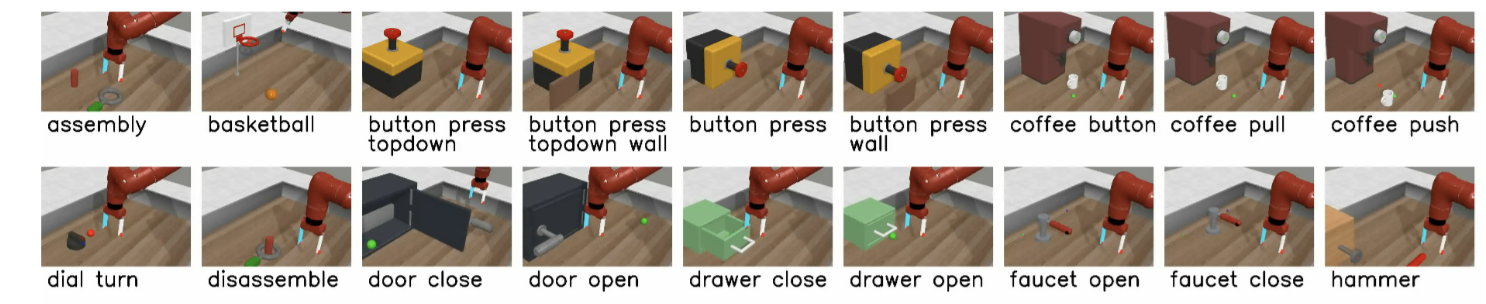}
					\caption{\textbf{Meta-World task visualizations.} Representative snapshots of the manipulation tasks used in our experiments (e.g., Assembly, Basketball, Button-Press variants, Coffee variants, Door variants, Drawer variants, Faucet variants, Hammer, etc.), illustrating the diverse objects and interaction dynamics in the Meta-World benchmark.}
					
					\label{meta}
				\end{figure*}

				The Meta-World benchmark~\cite{yu2020meta} comprises 50 diverse manipulation tasks that share a common set of dynamics. These tasks require a Sawyer robotic arm to interact with a variety of objects, each characterized by distinct geometric shapes, joint structures, and physical constraints. The challenge of this benchmark arises from the heterogeneity in state representations and reward functions across tasks, as the robot must achieve different objectives using task-specific interactions. The control input at each timestep is a 4-dimensional continuous action vector that governs the 3D position of the end-effector as well as the opening and closing of the gripper. We illustrate the simulation environments of selected tasks from Meta-World in Fig.~\ref{meta}.

				In the default Meta-World setup, goal positions are fixed, which can limit the realism and generalizability of learned policies. To better reflect more dynamic robotic scenarios, we adapt all tasks to a random-goal setting, denoted as MT50-rand. Evaluation is conducted using the average success rate across all 50 tasks, offering a holistic measure of policy generalization and performance in diverse conditions.
				
				To construct our offline datasets, we adopt the procedure introduced by, where individual SAC agents are trained per task until convergence. We collect 1 million transitions per task from each agent’s replay buffer, capturing the full trajectory of learning from random exploration to final convergence. This yields a large and varied dataset from which we derive two distinct versions:
				
				\begin{itemize}
					\item \textbf{Near-optimal} dataset: Aggregates the entire SAC training data (100M transitions in total), including both suboptimal and expert-level behaviors.
					\item \textbf{Sub-optimal} dataset: Extracted as the first 50\% of each task's SAC replay buffer (50M transitions total), representing primarily non-expert performance with reduced task mastery.
				\end{itemize}

				\subsection{Baselines}  
				We benchmark our proposed method, DaCe-DT, against a diverse suite of both offline and online multi-task learning baselines. The following provides a detailed description of the baselines used in this work:
				
				\paragraph{Offline Multi-Task Baselines.}
				\begin{enumerate}
					\item \textbf{MTBC}~\cite{hu2024harmodt}. This variant scales Behavior Cloning (BC) to the multi-task domain by introducing a task-ID conditioned policy network.
					
					\item \textbf{MTIQL}. Built upon the Implicit Q-Learning (IQL)~\cite{kostrikov2021offline} framework, MTIQL employs a multi-head critic structure and integrates task identity conditioning in the actor, offering a strong TD-based baseline to contrast against generative modeling approaches.
					
					\item \textbf{MTDIFF-P}~\cite{he2023diffusion}. A generative diffusion-based policy that leverages Transformer architectures and prompt-driven conditioning to perform multi-task offline planning and data modeling.
					
					\item \textbf{MTDT}. A multi-task adaptation of the Decision Transformer~\cite{chen2021decision}, where each task is encoded via a latent embedding $z$ concatenated with state inputs. At test time, the Transformer receives both reward-to-go and task embedding as inputs to produce tailored action predictions.
					
					\item \textbf{Prompt-DT}~\cite{xu2022prompting}. Based on the DT framework, Prompt-DT introduces trajectory prompt conditioning to enable better generalization across tasks and improve policy robustness to unseen task distributions.
					
					\item \textbf{HarmoDT}~\cite{hu2024harmodt}. Building upon the Prompt-DT framework, HarmoDT introduces a task-specific masking mechanism to mitigate gradient conflicts among tasks. HarmoDT comes in three variants: \textbf{HarmoDT-R}, which applies a random masking strategy for each task. \textbf{HarmoDT-M}, which measures parameter importance using the absolute value of gradients. and \textbf{HarmoDT-F}, which uses the squared value of gradients to quantify parameter importance~\cite{zhang2024learning}.
					
				\end{enumerate}
				
				\paragraph{Online Multi-Task Baselines.}
				In addition to the offline counterparts, we also include comparisons with representative online learning methods to provide a broader perspective on our approach’s effectiveness:
				
				\begin{enumerate}
					\item \textbf{CARE}~\cite{sodhani2021multi}. This approach leverages auxiliary metadata and multiple encoders to construct expressive task representations, supporting better transfer and generalization in multi-task learning.
					
					\item \textbf{PaCO}~\cite{sun2022paco}. Introduces a compositional parameter-sharing mechanism where task-specific parameters are dynamically recombined, promoting flexibility and efficiency across task distributions.
					
					\item \textbf{Soft-M}~\cite{yang2020multi}. Utilizes a modular routing network to softly blend multiple functional modules, enabling adaptive skill reuse across a range of tasks.
					
					\item \textbf{D2R}~\cite{he2024not}. Implements a dynamic routing strategy that selects different subsets of modules per task, thereby offering scalable and fine-grained control over model capacity based on task complexity.
					
					\item \textbf{MOORE}~\cite{hendawy2023multi}. Maintains a shared representation space across tasks using a Mixture of Orthogonal Experts, where orthogonality is enforced via a Gram-Schmidt process. This promotes diversity among expert representations and mitigates task interference, achieving state-of-the-art performance on Meta-World.
					
					\item \textbf{ARS-LoRA}~\cite{cho2025ars}. Introduces an adaptive reward scaling strategy that dynamically normalizes per-task reward magnitudes based on historical statistics, combined with a periodic network reset mechanism to prevent plasticity loss. ARS-LoRA further integrates Low-Rank Adaptation (LoRA) to enable parameter-efficient task-specific fine-tuning within the multi-task policy.
				\end{enumerate}
				\section{Hyper-parameters and Resources}
				\label{appendix:hyperparams}
				\subsection{Hyper-parameters}
				This section outlines the detailed experimental configurations used in our study. The policy model is built upon a Transformer architecture and implemented based on the open-source project minGPT. All experiments are conducted on the MetaWorld benchmark with a fixed random seed to ensure reproducibility. The model architecture and training hyperparameters are summarized in Table~\ref{SoCo}.
				
				\begin{table}[h]
					\caption{DaCe-DT training hyperparameters}
					\label{SoCo}
					\centering
					\begin{tabular}{lc}
                    
						\toprule
						\textbf{Parameter} & \textbf{Value} \\
						\midrule
						Number of Layers & 6 \\
						Number of Attention Heads & 8 \\
						Embedding Dimension & 256 \\
						Nonlinearity Function & RELU \\
						Context Length K & 20 \\
						Dropout & 0.1 \\
						Max Iterations & 100,000 \\
						Batch Size & 16 \\
						\bottomrule
					\end{tabular}
					
				\end{table}
				In addition, Tables~\ref{harmodt} and~\ref{prompt} provide the hyperparameter settings specific to each baseline method.
				\begin{table}[h]
					\caption{HarmoDT training hyperparameters}
					\label{harmodt}
					\centering
					\begin{tabular}{lc}
						\toprule
						\textbf{Parameter} & \textbf{Value} \\
						\midrule
						Sparsity & 0.2 \\
						Batch Size & 8 \\
						Mask Changing Interval & 5000 \\
						Minimum of Mask Changing$\eta_{min}$ & 0 \\
						Maximum of Mask Changing$\eta_{max}$ & 100 \\
						Maximum Threshold in Mask & 25 \\
						\bottomrule
					\end{tabular}
					
				\end{table}
				
				\begin{table}[h]
					\caption{Prompt-DT training hyperparameters}
					\label{prompt}
					\centering
					\begin{tabular}{lc}
						\toprule
						\textbf{Parameter} & \textbf{Value} \\
						\midrule
						K (length of context $\tau$) & 20 \\
						Batch Size & 8 \\
						Number of Evaluation Episodes for Each Task & 20 \\
						Learning Rate & 1e-4 \\
						Learning Rate Decay Weight & 1e-4 \\
						Number of Layers & 6 \\
						Number of Attention Heads & 8 \\
						Embedding Dimension & 128 \\
						Activation & RELU \\
						\bottomrule
					\end{tabular}
					
				\end{table}
				In our experiments, all stochastic components are controlled using fixed random seeds to ensure reproducibility. It is worth noting that the performance of the same method can vary across different randomly sampled environment combinations. As mentioned in the experimental section, we evaluate the method under three different random seeds: 123, 234, and 345, each corresponding to a distinct set of task combinations.

				\subsection{Computational Resources}  
				All experiments are conducted using NVIDIA 4090 GPUs. For each environment configuration—MT5, MT30, and MT50—we train the models for 1 million steps. On the near-optimal dataset, the training time is approximately 16 hours for MT5, 76 hours for MT30, and 112 hours for MT50. In contrast, on the suboptimal dataset, which contains fewer trajectories, the corresponding training times are reduced to 10 hours, 54 hours, and 84 hours, respectively. Each environment needs to be trained three times with different seeds.

				\section{Additional Experimental Analysis}

				\subsection{Maze2D Multi-Task Experiments}
				\label{appendix:maze2d}

				Maze2D~\cite{fu2020d4rl} is a sparse-reward navigation environment in which an agent must reach a goal position within a 2D maze. Unlike Meta-World, Maze2D provides no dense reward shaping, making it a challenging testbed for evaluating prompt quality and retrieval effectiveness without value-function calibration.

\begin{figure}[h]
					\centering
					\includegraphics[width=0.9\linewidth]{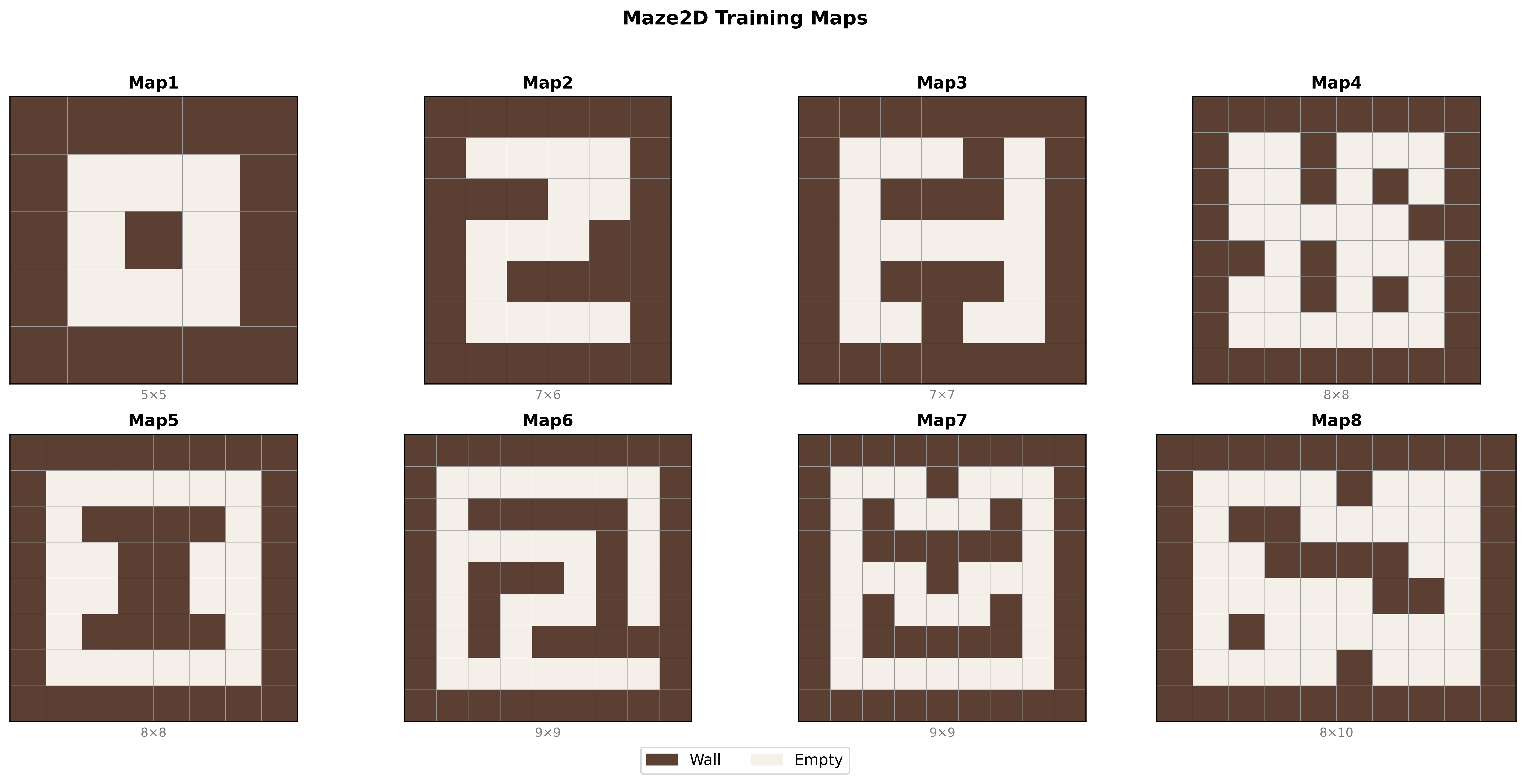}
					\caption{The 8 maze layouts used in our multi-task \textsc{Maze2D} benchmark. Layouts range from $5{\times}5$ to $8{\times}10$ grids and vary in wall topology, yielding 8 navigation tasks with distinct structural complexity.}
					\label{fig:maze2d}
				\end{figure}

				\textbf{Multi-Task Extension.}
				To construct a multi-task Maze2D benchmark, we create 8 distinct maze layouts by modifying the environment configuration files, as illustrated in Fig.~\ref{fig:maze2d}. Each layout constitutes a separate task with its own pre-collected offline dataset. Since VARC relies on dense reward signals, it is disabled in this experiment; only RAPC and LGPM are evaluated.

				\textbf{Results.}
				As shown in Table~\ref{tab:maze2d}, DaCe-DT achieves an average score of 36.9 compared to 18.2 for Prompt-DT, demonstrating that embedding-based retrieval and adaptive prompt length substantially improve performance even under entirely sparse rewards.

				\begin{table}[h]
					\centering
					\caption{Performance on sparse-reward \textsc{Maze2D} with 8 custom maze layouts. VARC is disabled; only RAPC and LGPM are used. Score is cumulative reward per episode.}
					\label{tab:maze2d}
					\small
					\begin{tabular}{lcc}
						\toprule
						Task & \textbf{Ours} & Prompt-DT \\
						\midrule
						maze2d-1 & \textbf{106.5} & 48.0 \\
						maze2d-2 & \textbf{95.1}  & 22.0 \\
						maze2d-3 & 0.0            & \textbf{12.0} \\
						maze2d-4 & \textbf{32.9}  & 7.5  \\
						maze2d-5 & 4.5            & \textbf{41.1} \\
						maze2d-6 & \textbf{22.6}  & 14.9 \\
						maze2d-7 & \textbf{33.2}  & 0.0  \\
						maze2d-8 & 0.0            & 0.0  \\
						\midrule
						Average  & \textbf{36.9}  & 18.2 \\
						\bottomrule
					\end{tabular}
				\end{table}

				\subsection{Per-Task Results on MetaWorld MT30}
				\label{appendix:mt30}

				Table~\ref{tab:mt30_pertask} reports the per-task episodic return and success rate of DaCe-DT on MetaWorld MT30 (near-optimal dataset, seed 123). The overall mean return is 3185.45 and mean success rate is 0.933.

				\begin{table}[h]
					\centering
					\caption{Per-task episodic return and success rate of DaCe-DT on MetaWorld MT30 (near-optimal, seed 123). Mean return: 3185.45; mean success: 0.933.}
					\label{tab:mt30_pertask}
					\scriptsize
					\setlength{\tabcolsep}{4pt}
					\begin{tabular}{lcc|lcc|lcc}
						\toprule
						\textbf{Task} & \textbf{Return} & \textbf{Succ.} &
						\textbf{Task} & \textbf{Return} & \textbf{Succ.} &
						\textbf{Task} & \textbf{Return} & \textbf{Succ.} \\
						\midrule
						basketball-v2           & 2198.46 & 1.0 & door-close-v2        & 4389.19 & 1.0 & handle-press-v2      & 4359.66 & 1.0 \\
						bin-picking-v2          & 4350.81 & 1.0 & door-lock-v2         & 3152.24 & 0.9 & handle-pull-side-v2  & 3646.42 & 1.0 \\
						button-press-topdown-v2 & 3669.39 & 1.0 & door-open-v2         & 3715.46 & 1.0 & handle-pull-v2       & 4264.39 & 1.0 \\
						button-press-v2         & 2608.85 & 1.0 & door-unlock-v2       & 3840.95 & 1.0 & lever-pull-v2        & 1880.16 & 1.0 \\
						button-press-wall-v2    & 2304.72 & 1.0 & hand-insert-v2       & 3591.48 & 1.0 & peg-insert-side-v2   & 4506.66 & 1.0 \\
						coffee-button-v2        & 1658.16 & 1.0 & drawer-close-v2      & 4831.95 & 1.0 & pick-place-wall-v2   & 3032.87 & 0.9 \\
						coffee-pull-v2          & 1299.40 & 1.0 & drawer-open-v2       & 3837.06 & 1.0 & pick-out-of-hole-v2  & 3382.60 & 0.8 \\
						coffee-push-v2          & 2789.15 & 1.0 & faucet-open-v2       & 4571.72 & 1.0 & reach-v2             & 2222.18 & 0.9 \\
						dial-turn-v2            & 3141.51 & 0.5 & faucet-close-v2      & 3706.75 & 1.0 & push-back-v2         & 2099.79 & 1.0 \\
						disassemble-v2          &  696.19 & 0.7 & handle-press-side-v2 & 4712.93 & 1.0 & push-v2              & 1102.48 & 0.3 \\
						\midrule
						\multicolumn{9}{l}{\textbf{Mean:} Return = 3185.45 \quad\quad Success = 0.9333} \\
						\bottomrule
					\end{tabular}
				\end{table}

				\subsection{Per-Task Results on MetaWorld MT50}
				\label{appendix:mt50_pertask}

				Table~\ref{tab:mt50_pertask} reports the per-task episodic return and success rate of DaCe-DT on MetaWorld MT50 (near-optimal dataset, seed 123). The overall mean return is 2983.47 and mean success rate is 0.884.

				\begin{table}[h]
					\centering
					\caption{Per-task episodic return and success rate of DaCe-DT on MetaWorld MT50 (near-optimal, seed 123). Mean return: 2983.47; mean success: 0.884.}
					\label{tab:mt50_pertask}
					\scriptsize
					\setlength{\tabcolsep}{4pt}
					\begin{tabular}{lcc|lcc|lcc}
						\toprule
						\textbf{Task} & \textbf{Return} & \textbf{Succ.} &
						\textbf{Task} & \textbf{Return} & \textbf{Succ.} &
						\textbf{Task} & \textbf{Return} & \textbf{Succ.} \\
						\midrule
						basketball-v2              & 2337.14 & 1.0 & faucet-open-v2          & 4574.86 & 1.0 & plate-slide-back-side-v2     &  652.54 & 1.0 \\
						bin-picking-v2             & 3958.31 & 1.0 & faucet-close-v2         & 4663.15 & 1.0 & soccer-v2                    & 1043.56 & 0.3 \\
						button-press-topdown-v2    & 3482.01 & 0.9 & handle-press-side-v2    & 4731.69 & 1.0 & push-wall-v2                 & 3789.07 & 0.8 \\
						button-press-v2            & 2396.76 & 1.0 & handle-press-v2         & 4315.96 & 0.9 & shelf-place-v2               & 1633.87 & 0.9 \\
						button-press-wall-v2       & 2155.54 & 1.0 & handle-pull-side-v2     & 3092.48 & 1.0 & sweep-into-v2                & 4176.40 & 0.9 \\
						coffee-button-v2           & 1695.16 & 0.9 & handle-pull-v2          & 4424.69 & 1.0 & sweep-v2                     & 2449.86 & 0.7 \\
						coffee-pull-v2             & 1231.68 & 0.8 & lever-pull-v2           & 1336.97 & 1.0 & window-open-v2               & 1347.92 & 1.0 \\
						coffee-push-v2             & 1001.34 & 0.5 & peg-insert-side-v2      & 3855.10 & 1.0 & window-close-v2              & 1612.94 & 1.0 \\
						dial-turn-v2               & 3031.78 & 0.4 & pick-place-wall-v2      & 1973.96 & 0.9 & assembly-v2                  & 3318.62 & 1.0 \\
						disassemble-v2             &  325.36 & 0.4 & pick-out-of-hole-v2     & 3037.74 & 1.0 & button-press-topdown-wall-v2 & 3070.91 & 1.0 \\
						door-close-v2              & 4385.56 & 1.0 & reach-v2                & 2998.81 & 0.9 & hammer-v2                    & 4298.86 & 1.0 \\
						door-lock-v2               & 3293.90 & 0.8 & push-back-v2            & 4149.67 & 1.0 & peg-unplug-side-v2           & 2150.17 & 1.0 \\
						door-open-v2               & 3182.73 & 1.0 & push-v2                 & 1880.48 & 0.5 & reach-wall-v2                & 3004.80 & 0.7 \\
						door-unlock-v2             & 3851.01 & 1.0 & pick-place-v2           & 2405.79 & 0.9 & stick-push-v2                & 3352.03 & 1.0 \\
						hand-insert-v2             & 3438.67 & 0.9 & plate-slide-v2          & 4592.15 & 1.0 & stick-pull-v2                & 3589.41 & 0.2 \\
						drawer-close-v2            & 4826.31 & 1.0 & plate-slide-side-v2     & 3300.93 & 1.0 & box-close-v2                 & 3786.98 & 1.0 \\
						drawer-open-v2             & 3757.75 & 1.0 & plate-slide-back-v2     & 2210.22 & 1.0 &                              &         &     \\
						\midrule
						\multicolumn{9}{l}{\textbf{Mean:} Return = 2983.47 \quad\quad Success = 0.8840} \\
						\bottomrule
					\end{tabular}
				\end{table}

	\begin{table}[h]
	\centering
	\caption{Prompt lengths learned by LGPM per task under different configurations. The same task converges to different lengths depending on the task combination, reflecting LGPM's adaptivity to task heterogeneity.}
	\label{tab:lgpm_lengths}
	\small
	\begin{tabular}{lccc}
		\toprule
		Task & MT5 & MT30 (near-opt.) & MT30 (sub-opt.) \\
		\midrule
		basketball-v2             & 18 & 24 & 36 \\
		reach-v2                  & 8  & 6  & 10 \\
		button-press-topdown-v2   & 10 & 20 & 26 \\
		button-press-v2           & 20 & 10 & 19 \\
		button-press-wall-v2      & 10 & 21 & 27 \\
		\bottomrule
	\end{tabular}
\end{table}
				\subsection{Learned Prompt Lengths (LGPM)}
				\label{appendix:lgpm_lengths}

				Table~\ref{tab:lgpm_lengths} reports the prompt lengths learned by LGPM for representative tasks across different environment configurations. Notably, the same task may converge to different prompt lengths under different task combinations, reflecting how co-trained task complexity influences the optimal context window for each individual task.

	\begin{figure*}[t]
	\begin{center}
		\centerline
		{\includegraphics[width=\columnwidth]{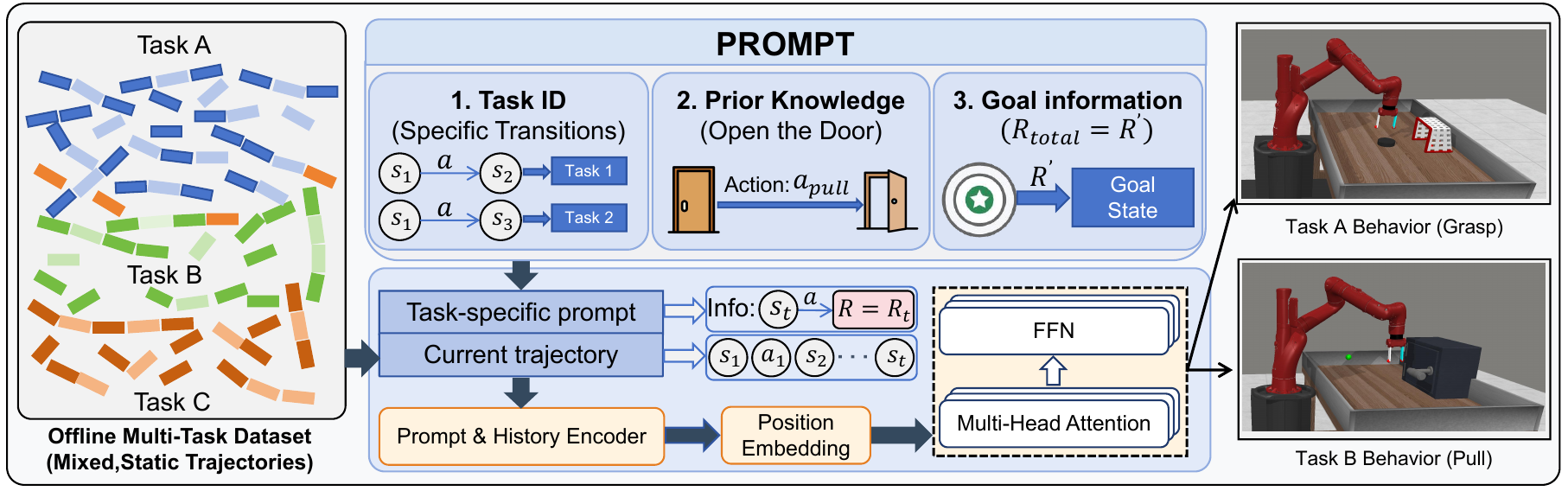}}
		\caption{
			Overview of prompt-based offline multi-task learning. The prompt disambiguates mixed offline trajectories by encoding task identity, behavioral priors, and goal/return targets, guiding the Transformer to produce task-specific behaviors (e.g., Grasp vs. Pull).
		}
		\label{fig:prompt}
	\end{center}
\end{figure*}
				\section{Related Work}
				
				\subsection{Offline Reinforcement Learning}
				Offline RL~\cite{levine2020offline} aims to learn effective decision-making policies solely from pre-collected static datasets, without the need for further interaction with the environment. This setting is particularly valuable in high-stakes or resource-constrained domains~\cite{kumar2021workflow,shi2021offline,ghasemi2025personalized}, where data collection is costly, risky, or impractical. A central challenge in this domain arises from the discrepancy between the distribution of the training data and the learned policy—commonly referred to as distributional shift—which can severely degrade performance. To address this issue, various approaches have proposed conservative policy updates or regularization techniques that keep the learned policy close to the behavior policy~\cite{kostrikov2021offline}.
				A promising line of work in offline RL leverages conditional sequence modeling~\cite{chen2021decision}, where the agent is trained to predict future actions conditioned on historical trajectories composed of state-action-reward sequences. These methods reformulate policy learning as a sequence prediction task, aligning naturally with supervised learning frameworks and implicitly constraining the learned behavior to remain within the support of the dataset. This class of methods includes the popular Decision Transformer and its extensions, which have demonstrated strong performance across various benchmarks.
				
				In parallel, recent advances have explored the use of diffusion models for offline RL, inspired by their success in generative modeling. These approaches model either the policy or environment dynamics via iterative denoising processes, enabling more expressive action distributions. Methods such as Diffuser and its variants have shown impressive results across a range of tasks.

				\subsection{Multi-task Reinforcement Learning}
				Multi-task reinforcement learning (MTRL) focuses on training a unified policy that can generalize across a range of distinct tasks. Numerous approaches have been proposed in recent years to address this challenge~\cite{yang2020multi,sodhani2021multi}. A common formulation treats the multi-task problem as a conditional learning problem, where task identifiers or goals are provided as context inputs, following the design pattern of goal-conditioned RL~\cite{he2023diffusion,xu2022prompting}. Another line of research focuses on identifying optimal parameter subspaces for each task, utilizing routing mechanisms or weight-sharing strategies to compute task-specific weighted combinations of network layers~\cite{hu2024harmodt,sun2022paco}.
				
				Building upon these foundations, offline multi-task reinforcement learning (offline MTRL) combines the principles of offline reinforcement learning and multi-task learning, leveraging task-mixed datasets to enable knowledge transfer without environment interaction. This approach is particularly valuable in scenarios where online data collection is expensive or risky, allowing agents to learn from pre-collected datasets spanning multiple tasks.
				
				However, a persistent challenge in both online and offline MTRL is the presence of conflicting gradients among tasks, often referred to as negative transfer or gradient interference~\cite{shi2023recon,tang2023concrete,d2024sharing,lee2022multi,caruana1997multitask}. This problem arises when task and data heterogeneity induce gradient conflicts that hinder effective parameter sharing, particularly in offline settings where data distribution varies significantly across tasks.
				
				To address these gradient conflicts in offline MTRL, existing approaches~\cite{hu2024harmodt,sun2022paco,he2023diffusion} employ routing mechanisms or masking mechanisms to mitigate the adverse effects of gradient interference. Parameter-level methods adopt task-specific binary mask matrices to determine optimal subspaces within shared parameters, while routing-based approaches dynamically allocate network capacity across different tasks. Although these methods have demonstrated some success, gradient interference continues to be a fundamental challenge that requires resolution in offline multi-task reinforcement learning.
				
				\subsection{Prompt-DT}
				\label{promptdt}
				Although DT is effective in single-task offline RL, directly applying it to multi-task datasets is non-trivial because trajectories from different tasks are often heterogeneous in dynamics, reward scales, and optimal behaviors.
				Prompt-DT~\cite{xu2022prompting} extends DT to multi-task learning by introducing a \emph{prompting} mechanism that provides task-specific context.
				For each task $T_i$, Prompt-DT samples a short $K$-step demonstration subtrajectory (a prompt) from the target task and prepends it to the current rollout, yielding
				\begin{align}
					\tau^{\text{input}}_{i,t} = (&\widehat{R}^*_{i,1}, s^*_{i,1}, a^*_{i,1}, \dots, \widehat{R}^*_{i,K}, s^*_{i,K}, a^*_{i,K}, \nonumber\\
					&\widehat{R}_{i,t-K+1}, s_{i,t-K+1}, a_{i,t-K+1}, \dots, \widehat{R}_{i,t}, s_{i,t}, a_{i,t}).
				\end{align}

                \section{Limitations}

DaCe-DT addresses key data-centric challenges in offline multi-task reinforcement
learning, yet several broader limitations warrant discussion.
First, as an offline method, DaCe-DT is fundamentally bounded by the quality and
diversity of the pre-collected dataset; it cannot actively explore the environment
to recover from distributional gaps, which may hinder performance in domains where
high-quality offline data is scarce or difficult to obtain.
Second, our evaluation is confined to simulation benchmarks with well-defined task
boundaries and reward signals. Deploying DaCe-DT in open-world or real-robot
settings, where observations are noisy, tasks are ambiguous, and reward feedback
is sparse or absent, remains an important open challenge.

				
				\newpage
				\section*{NeurIPS Paper Checklist}

The checklist is designed to encourage best practices for responsible machine learning research, addressing issues of reproducibility, transparency, research ethics, and societal impact. Do not remove the checklist: {\bf The papers not including the checklist will be desk rejected.} The checklist should follow the references and follow the (optional) supplemental material.  The checklist does NOT count towards the page
limit.

Please read the checklist guidelines carefully for information on how to answer these questions. For each question in the checklist:
\begin{itemize}
    \item You should answer \answerYes{}, \answerNo{}, or \answerNA{}.
    \item \answerNA{} means either that the question is Not Applicable for that particular paper or the relevant information is Not Available.
    \item Please provide a short (1--2 sentence) justification right after your answer (even for \answerNA).
\end{itemize}

{\bf The checklist answers are an integral part of your paper submission.} They are visible to the reviewers, area chairs, senior area chairs, and ethics reviewers. You will also be asked to include it (after eventual revisions) with the final version of your paper, and its final version will be published with the paper.

The reviewers of your paper will be asked to use the checklist as one of the factors in their evaluation. While \answerYes{} is generally preferable to \answerNo{}, it is perfectly acceptable to answer \answerNo{} provided a proper justification is given (e.g., error bars are not reported because it would be too computationally expensive'' or ``we were unable to find the license for the dataset we used''). In general, answering \answerNo{} or \answerNA{} is not grounds for rejection. While the questions are phrased in a binary way, we acknowledge that the true answer is often more nuanced, so please just use your best judgment and write a justification to elaborate. All supporting evidence can appear either in the main paper or the supplemental material, provided in appendix. If you answer \answerYes{} to a question, in the justification please point to the section(s) where related material for the question can be found.

IMPORTANT, please:
\begin{itemize}
    \item {\bf Delete this instruction block, but keep the section heading ``NeurIPS Paper Checklist"},
    \item  {\bf Keep the checklist subsection headings, questions/answers and guidelines below.}
    \item {\bf Do not modify the questions and only use the provided macros for your answers}.
\end{itemize}


\begin{enumerate}

\item {\bf Claims}
    \item[] Question: Do the main claims made in the abstract and introduction accurately reflect the paper's contributions and scope?
    \item[] Answer: \answerYes{}
    \item[] Justification: The abstract and introduction clearly state the three data-centric bottlenecks and the corresponding mechanisms (LGPM, RAPC, VARC), which are validated by experimental results in \S\ref{sec:experiments}.

\item {\bf Limitations}
    \item[] Question: Does the paper discuss the limitations of the work performed by the authors?
    \item[] Answer: \answerYes{} 
    \item[] Justification: We discuss limitations in a dedicated paragraph in the appendix,
covering the dependence on offline data quality and the sim-to-real generalization gap.

\item {\bf Theory assumptions and proofs}
    \item[] Question: For each theoretical result, does the paper provide the full set of assumptions and a complete (and correct) proof?
    \item[] Answer: \answerNA{}
    \item[] Justification: This paper does not include formal theorems or proofs; all contributions are empirical.

    \item {\bf Experimental result reproducibility}
    \item[] Question: Does the paper fully disclose all the information needed to reproduce the main experimental results of the paper to the extent that it affects the main claims and/or conclusions of the paper (regardless of whether the code and data are provided or not)?
    \item[] Answer: \answerYes{}
    \item[] Justification: Appendix~B provides full hyperparameter settings and compute details. Experiments use fixed random seeds (123, 234, 345) on publicly available benchmarks (Meta-World, D4RL).

\item {\bf Open access to data and code}
    \item[] Question: Does the paper provide open access to the data and code, with sufficient instructions to faithfully reproduce the main experimental results, as described in supplemental material?
    \item[] Answer: \answerNo{}
    \item[] Justification: Code will be released upon paper acceptance. At submission time, the repository is withheld to preserve double-blind anonymity.

\item {\bf Experimental setting/details}
    \item[] Question: Does the paper specify all the training and test details (e.g., data splits, hyperparameters, how they were chosen, type of optimizer) necessary to understand the results?
    \item[] Answer: \answerYes{}
    \item[] Justification: Dataset construction, evaluation protocol, hyperparameters, and optimizer details are described in \S\ref{sec:experiments} and Appendix~B.

\item {\bf Experiment statistical significance}
    \item[] Question: Does the paper report error bars suitably and correctly defined or other appropriate information about the statistical significance of the experiments?
    \item[] Answer: \answerYes{}
    \item[] Justification: All main results are averaged over three random seeds; standard deviations are reported in all tables (Tables~\ref{tb:scale} and~\ref{tb:50task}).

\item {\bf Experiments compute resources}
    \item[] Question: For each experiment, does the paper provide sufficient information on the computer resources (type of compute workers, memory, time of execution) needed to reproduce the experiments?
    \item[] Answer: \answerYes{}
    \item[] Justification: Appendix~B reports that all experiments are run on NVIDIA 4090 GPUs, with training times of approximately 16h (MT5), 76h (MT30), and 112h (MT50) on the near-optimal dataset.

\item {\bf Code of ethics}
    \item[] Question: Does the research conducted in the paper conform, in every respect, with the NeurIPS Code of Ethics \url{https://neurips.cc/public/EthicsGuidelines}?
    \item[] Answer: \answerYes{}
    \item[] Justification: This research uses publicly available offline RL benchmarks and does not involve human subjects, sensitive data, or harmful applications.

\item {\bf Broader impacts}
    \item[] Question: Does the paper discuss both potential positive societal impacts and negative societal impacts of the work performed?
    \item[] Answer: \answerNA{}
    \item[] Justification: This work focuses on foundational offline multi-task RL methodology applied to robotic manipulation benchmarks. We do not foresee direct negative societal impacts.

\item {\bf Safeguards}
    \item[] Question: Does the paper describe safeguards that have been put in place for responsible release of data or models that have a high risk for misuse (e.g., pre-trained language models, image generators, or scraped datasets)?
    \item[] Answer: \answerNA{}
    \item[] Justification: This paper does not release pre-trained language models, image generators, or internet-scraped datasets, and thus poses no such risks.

\item {\bf Licenses for existing assets}
    \item[] Question: Are the creators or original owners of assets (e.g., code, data, models), used in the paper, properly credited and are the license and terms of use explicitly mentioned and properly respected?
    \item[] Answer: \answerYes{}
    \item[] Justification: All benchmarks and code packages used (Meta-World, D4RL, Decision Transformer, HarmoDT) are properly cited. These are publicly available research artifacts.

\item {\bf New assets}
    \item[] Question: Are new assets introduced in the paper well documented and is the documentation provided alongside the assets?
    \item[] Answer: \answerNA{}
    \item[] Justification: This paper does not release new datasets or model checkpoints as standalone assets.

\item {\bf Crowdsourcing and research with human subjects}
    \item[] Question: For crowdsourcing experiments and research with human subjects, does the paper include the full text of instructions given to participants and screenshots, if applicable, as well as details about compensation (if any)?
    \item[] Answer: \answerNA{}
    \item[] Justification: This paper does not involve crowdsourcing or research with human subjects.

\item {\bf Institutional review board (IRB) approvals or equivalent for research with human subjects}
    \item[] Question: Does the paper describe potential risks incurred by study participants, whether such risks were disclosed to the subjects, and whether Institutional Review Board (IRB) approvals (or an equivalent approval/review based on the requirements of your country or institution) were obtained?
    \item[] Answer: \answerNA{}
    \item[] Justification: No human subjects are involved in this research; IRB approval is not required.

\item {\bf Declaration of LLM usage}
    \item[] Question: Does the paper describe the usage of LLMs if it is an important, original, or non-standard component of the core methods in this research? Note that if the LLM is used only for writing, editing, or formatting purposes and does \emph{not} impact the core methodology, scientific rigor, or originality of the research, declaration is not required.
    \item[] Answer: \answerNA{}
    \item[] Justification: LLMs are not a core component of the proposed method. Any LLM use was limited to writing assistance and does not affect the scientific methodology.

\end{enumerate}

			\end{document}